%% file: main.tex
\documentclass[11pt,a4paper]{article}
\usepackage[preprint]{acl}
\usepackage{times}
\usepackage{latexsym}
\usepackage[T1]{fontenc}
\usepackage[utf8]{inputenc}
\usepackage{microtype}
\usepackage{courier}
\usepackage{amsmath,amssymb,booktabs,graphicx,multirow,tikz}
\usetikzlibrary{arrows.meta,positioning}
\input{stats/paper_statistics.tex}
\title{Where Did the Evidence Go?\\
Diagnosing Proof-Set Loss in Long-Term Conversational Memory}
\author{\normalfont%
  \begin{tabular}{@{}c@{\qquad}c@{}}
    \textbf{Bogdan Roshchupkin}\textsuperscript{1} &
    \textbf{Maria Khodorchenko}\textsuperscript{2} \\
    {\texttt{rosupkinbogdan6@gmail.com}} &
    {\texttt{mkhodorchenko@itmo.ru}}
  \end{tabular} \\[0.5em]
  \textsuperscript{1}Sber \qquad
  \textsuperscript{2}ITMO University
}
\hypersetup{
  pdftitle={Where Did the Evidence Go? Diagnosing Proof-Set Loss in Long-Term Conversational Memory},
  pdfauthor={Bogdan Roshchupkin; Maria Khodorchenko},
  pdfkeywords={conversational memory, evidence retrieval, retrieval-augmented generation}
}
\begin{document}
\maketitle
\begin{abstract}
Long-term conversational question answering can require several separated
dialogue turns that support an answer only
together.  We introduce EviCover, a retrieval approach designed to retain this
complementary evidence.  It converts questions into observable evidence
requirements, checks requirement coverage using retrieved source identifiers,
and selects turns jointly.  We track annotated support sets - the
benchmark-labeled dialogue turns associated with each question - through
retrieval, reranking, and final answer-generation context.  We evaluate the
baseline memory pipeline on generated, human-verified LoCoMo conversations and
human-written REALTALK conversations, and examine EviCover on REALTALK's 108
multi-hop questions.  Compared with the hybrid lexical/dense baseline,
support-turn recall among 60 candidate turns rises from \BaselinePoolRecall{}
to \RequirementPoolMatchedRecall{}, and recall in the final ten-turn context
rises from \BaselineFinalRecall{} to \SetwiseFinalRecall{}.
The evidence funnel identifies separate discovery and selection failures,
while paired answer evaluation shows that greater evidence recall does not
yet establish a correctness gain.  This analysis locates where complementary
evidence is lost and distinguishes evidence retrieval from its use in answer
generation.
\end{abstract}

\section{Introduction}
To answer questions about earlier interactions, long-term assistants retrieve
preferences, relationships, plans, and events from conversations distributed
across sessions.  A dialogue turn is an individual message from a participant.
Recent benchmarks now test not
only factual recall but also temporal reasoning, long-range understanding, and
implicit constraints \citep{maharana2024locomo,lee2025realtalk,
wu2025longmemeval,hu2026memoryagentbench,li2026locomoplus}.  These settings make
the memory problem different from retrieving one independently relevant
passage.  A question such as ``Does this person behave irresponsibly?'' may be
supported by one episode about health precautions and another about work
obligations.  Either episode is relevant, but neither alone justifies the
conclusion.

Retrieval metrics such as mean reciprocal rank measure how early a supporting
turn appears in a ranking.  They do not show whether all complementary turns
needed to answer a question survive reranking and enter the final context.
Here, the \emph{reader} is the language model that generates an answer from
the selected dialogue turns.  A turn can be retrieved initially and then
discarded when candidates are scored independently or the context is truncated.
Conversely, a retrieved turn may support an answer without being included in
the benchmark annotations.  We study these retrieval stages through an
\emph{annotated proof set}: all benchmark-labeled dialogue turns associated
with a question.  
We evaluate a provenance-preserving memory pipeline on two types of conversation
benchmarks: generated, human-verified LoCoMo and human-written REALTALK.  The baseline
configuration is selected using LoCoMo and held fixed for REALTALK evaluation.
Multi-hop questions form the weakest REALTALK slice.  For these questions,
EviCover retrieves evidence for explicit requirements and jointly selects
complementary turns.  Figure~\ref{fig:method} gives an overview of EviCover.
We then compare retrieval
trajectories, memory representations, and final-set selection using the same
lexical and dense indexes, raw-turn source identifiers, and answer-generation
model and prompts.  We address two research questions:

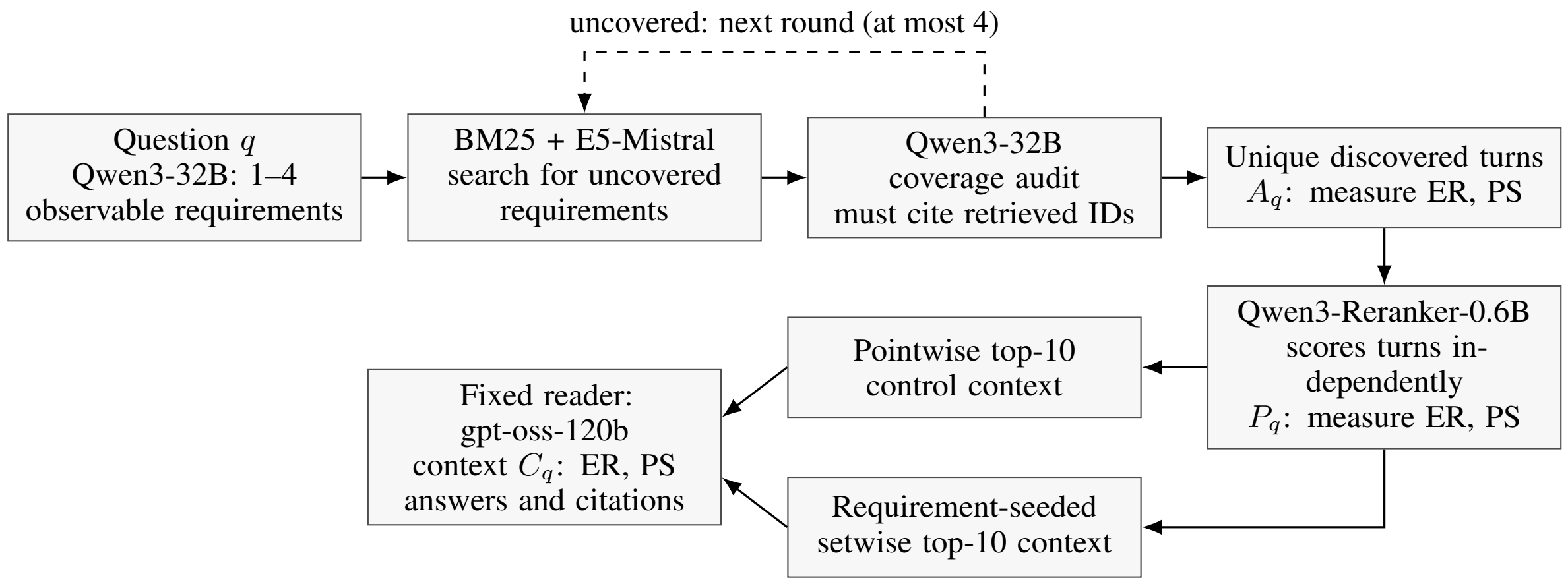
\begin{figure*}[t]
  \centering
  \resizebox{0.98\textwidth}{!}{\input{figures/method_overview.tex}}
  \caption{EviCover overview: provenance-constrained proof-set retrieval.  Requirement coverage is
  accepted only with a retrieved turn identifier.  The pointwise and setwise
  top-10 contexts use the same discovered candidates and relevance scores; only
  final set selection changes.  ER and PS are measured at every stored boundary.}
  \label{fig:method}
\end{figure*}

\begin{itemize}
  \item \textbf{RQ1:} How does the baseline perform on generated and human-written
  conversations, and at which retrieval stages are annotated proof sets lost?
  \item \textbf{RQ2:} Do requirement-guided discovery and setwise selection
  improve evidence delivery and answers enough to justify their cost?
\end{itemize}

We make two contributions.  (1) We formalize an evidence funnel that reports
support-turn recall and complete-proof-set survival at discovery, reranking, and
reader-context boundaries, and augment it with turn-, session-, day-, and
lexical-distance diagnostics.  (2) We conduct a controlled
evaluation of EviCover on REALTALK - observable requirement discovery,
provenance-checked coverage, and set-level selection - alongside baseline results
on LoCoMo, and jointly report evidence gains over flat hybrid retrieval and a
pointwise selection control, answer evaluation, and inference cost.

\section{Related Work}
Prior work addresses how memories are organized, how supporting information is
retrieved, and how memory is evaluated.  We review these choices in that order.

\textbf{Memory organization and recall.}  Structured memories aim to make related
information easier to recover.  A-MEM builds dynamic note networks
\citep{xu2025amem}; HippoRAG~2 integrates passages with graph-based associative
retrieval \citep{gutierrez2025hipporag2}; and Associa combines an event-centric
graph with iterative recall \citep{zhang2025associa}.  For document question
answering, KG-augmented RAG combines reusable graph-community summaries with
textual evidence \citep{karkout2026efficient}.  APEX-MEM adds temporal
structure and retrieval-time conflict resolution \citep{banerjee2026apexmem},
while MemGAS associates and selects memories at multiple granularities
\citep{xu2026memgas}.  Memory-R1 learns memory operations and answer-side
preselection \citep{yan2026memoryr1}, and MemCoRL alternates optimization of
retrieval and utilization \citep{liu2026memcorl}.  These approaches motivate
richer discovery and memory use.  We examine whether complementary evidence
found during discovery survives selection for the final context.

\textbf{Multi-step retrieval and set selection.}  IRCoT interleaves retrieval
with intermediate reasoning \citep{trivedi2023ircot}, while StepChain GraphRAG
constructs query-local evidence paths \citep{ni2025stepchain}.  Both motivate
searching beyond the initial question.  ColBERT's token-level late interaction
provides finer relevance matching than one vector per turn
\citep{khattab2020colbert}, but relevance scoring and joint coverage are distinct
objectives.  Setwise ranking compares document groups \citep{zhuang2024setwise},
and setwise insertion edits an existing ranking
\citep{podolak2025setwiseinsertion}.  Most directly, SetR derives query
information requirements and selects passages that cover them jointly
\citep{lee2025setr}.  We apply this coverage objective to long-term conversations,
where supporting messages may occur in different sessions.  Coverage decisions
must cite retrieved turn identifiers, and the evidence funnel tracks losses
during discovery, independent scoring, and final context selection.

\textbf{Evaluation of long-term memory.}  Benchmarks expose different demands
on conversational memory.  LoCoMo tests temporal and causal QA on generated,
human-verified conversations \citep{maharana2024locomo}; REALTALK uses messages
exchanged by real people over 21 days \citep{lee2025realtalk}.  This contrast
motivates evaluating our baseline on both datasets.  LongMemEval analyzes
indexing, retrieval, and reading \citep{wu2025longmemeval}, while MemoryAgentBench
tests retrieval, test-time learning, long-range understanding, and selective
forgetting \citep{hu2026memoryagentbench}.  LoCoMo-Plus targets implicit
constraints under cue--trigger semantic disconnect, which surface-form metrics
can miss \citep{li2026locomoplus}.  Our evidence funnel complements this
evaluation by measuring whether the full annotated support set survives each
retrieval stage.

\section{Task and Evaluation}
\subsection{Problem setting}
For user or conversation $u$, let
$H_u=(t_1,\ldots,t_n)$ be the ordered raw dialogue turns.  Each $t_i$ is a whole
participant message with a source identifier.  A benchmark question $q$ has
reference answer $a_q^*$ and an annotated support set $G_q\subseteq H_u$.
Each item in $G_q$ is a whole turn labeled as supporting $q$; we call this
collection the annotated proof set.  A retrieval pipeline collects turns in
$A_q$ across search rounds, keeping one copy per source identifier.  It scores
these turns independently against $q$ and retains a ranked candidate set
$P_q\subseteq A_q$.  The final context $C_q\subseteq P_q$, with $|C_q|\le k$,
contains the selected turns supplied to the reader for answer generation.
The reader returns answer $y_q$ and cited source identifiers $Z_q$.
For the set operations below, turns are represented by their source identifiers.

At any stage $S\in\{A,P,C\}$, we report micro-averaged gold-turn evidence recall
\begin{equation}
  \mathrm{ER}(S)=\frac{\sum_q |G_q\cap S_q|}{\sum_q |G_q|},
  \label{eq:er}
\end{equation}
and macro-averaged complete-proof-set rate
\begin{equation}
  \mathrm{PS}(S)=\frac{1}{|Q|}\sum_q \mathbf{1}[G_q\subseteq S_q].
  \label{eq:ps}
\end{equation}
ER is the fraction of annotated support turns retained across questions; PS is
the fraction of questions for which every annotated supporting turn is retained.

\subsection{Retrieval and answer metrics}
\label{sec:metrics}
Recall@5 is the macro-average of $|G_q\cap C_q^{(5)}|/|G_q|$, where
$C_q^{(5)}$ contains the first five turns in the ranked final context.
Mean reciprocal rank (MRR) averages the reciprocal rank of the first retrieved
member of $G_q$ in the final context, with zero if none is retrieved.
Under our reporting convention,
a question with no annotated support contributes zero to both metrics.
Unlike ER, Recall@5 is macro-averaged per question and truncated at five.

For answers, normalization lowercases text, collapses whitespace, and removes
punctuation.  Exact match (EM) is equality after this normalization.  Token F1
is the harmonic mean of multiset-token precision and recall against $a_q^*$.
We use both as lexical agreement diagnostics, not sufficient measures of semantic
correctness.  LoCoMo and REALTALK report partial-match F1, while LongMemEval uses
LLM-based evaluation to accommodate equivalent answer formulations
\citep{maharana2024locomo,lee2025realtalk,wu2025longmemeval}.
Our annotation-based abstention correctness is one when the system abstains
exactly when $G_q=\varnothing$; missing annotations alone do not establish that
a question is unanswerable.  A citation is a dialogue-turn source identifier
returned with the answer to indicate its supporting evidence.
Citation validity requires $Z_q\ne\varnothing$ and $Z_q\subseteq C_q$.
Per-question citation precision and recall measure identifier overlap with
the annotated support set:
\begin{equation}
 \mathrm{CP}_q=\frac{|Z_q\cap G_q|}{|Z_q|},\qquad
 \mathrm{CR}_q=\frac{|Z_q\cap G_q|}{|G_q|},
\end{equation}
with zero for an empty denominator.  We macro-average these per-question scores
and binary indicators over the evaluated questions.  Our mechanically
\emph{citation-grounded answer} indicator is one only when $y_q$ is nonempty and
not an abstention, all citations are valid, and $Z_q\cap G_q\ne\varnothing$.
This is a provenance check, not a semantic entailment claim.

Lexical metrics penalize equivalents such as ``twice'' and ``2 times.''  We
therefore complement them with LLM-as-a-judge evaluation using Qwen3-32B,
which sees the question, reference, generated answer, retrieved context, and
source identifiers.  It assigns
\texttt{correct}, \texttt{partially correct}, \texttt{incorrect},
\texttt{correct abstention}, or \texttt{wrong abstention}, and separately marks
whether the answer is supported by the shown context.  We report
\emph{judge accepted} as the combined rate of correct answers and
judge-accepted abstentions, and report those labels separately when comparing
reader prompts.  An accepted abstention is a judge decision, not proof that the
full conversation is unanswerable.  Following known judge
bias concerns \citep{zheng2023judge}, these labels complement rather than replace
the deterministic metrics.  All principal intervals use 20,000 paired
non-parametric bootstrap resamples of the ten conversations.

\section{Method}
\subsection{Overview}
EviCover changes the multi-hop branch of the frozen retrieval pipeline.
It has three explicit stages.  First, a planner converts the question into up
to four observable evidence requirements and retrieves for each uncovered
requirement.  Second, a provenance audit may mark a requirement covered only
by citing an identifier from the retrieved candidates.  Third, the same
pointwise relevance model used by the baseline supplies a prior, while a
setwise selector builds ten complementary turns rather than returning the ten
highest independent scores.  Figure~\ref{fig:method} shows the data passed
between stages; gold support identifiers enter only after generation for
evaluation.

\subsection{Baseline retrieval pipeline}
The memory store preserves each raw turn and its source identifier.  BM25 is the
lexical channel: it ranks turns by exact query-term evidence with document-length
and term-frequency normalization \citep{robertson2009bm25}.  The semantic channel
embeds questions and turns into 4,096-dimensional vectors with
E5-Mistral-7B-Instruct, a text-embedding model trained to place semantically
related queries and passages near one another \citep{wang2024e5mistral}.  We use
cosine similarity between vectors for dense retrieval; ``E5-Mistral''
below denotes this concrete model.

At write time, each turn is linked to at most ten active turns that share an
entity identifier or an identical nonempty subject; matching predicates,
subjects, and entity-only matches yield \textsc{accumulates}, \textsc{extends},
and \textsc{specifies} edges, respectively.  For factual and temporal questions,
the ten strongest lexical/dense seeds start a depth-two walk over these edges
(and temporal \textsc{transforms} edges), with at most 12 returned neighbors per
seed.  For multi-hop questions, deterministic text rules produce at most six
query facets and add dialogue neighbors within $\pm2$ turns.
The benchmark-provided query type chooses the branch.  Weighted reciprocal-rank
fusion (RRF) combines lexical, dense, and graph rankings as
\begin{equation}
 s_{\mathrm{RRF}}(t)=\sum_{r}\frac{w_r}{60+\mathrm{rank}_r(t)},
\end{equation}
with $w_{\mathrm{BM25}}=0.25$, $w_{\mathrm{E5}}=4.0$, and
$w_{\mathrm{graph}}=1.5$ when the graph channel is present
\citep{cormack2009rrf}.  The best 60 fused turns are then scored independently by
the 0.6B Qwen3-Reranker model.  It receives one pair $(q,t)$ at a time and
converts its final yes/no logits to a relevance probability
\citep{zhang2025qwen3embedding}.  Because no candidate is scored in the context
of another candidate, we call this a \emph{pointwise relevance scorer}; it is
the only learned relevance model in the controlled baseline and
requirement-guided comparisons.

\subsection{Provenance-constrained requirement discovery}
Requirement-guided discovery starts with a Qwen3-32B planner that maps $q$ to
one to four observable, independently verifiable requirements
$R_q=\{r_1,\ldots,r_m\}$.  For example, an evaluative question about
irresponsibility can yield one requirement for behavior affecting others and
one for an unfulfilled obligation; it does not search for the abstract adjective
alone.  Each round retrieves up to 60 BM25 and 60 E5-Mistral results for one
uncovered requirement and merges them by RRF.  Candidates remain in $A_q$
across at most four rounds.

The planner then audits at most 16 shown candidates.  A requirement may be
marked covered only with a source identifier in $A_q$; code rejects absent or
malformed identifiers.  An uncovered requirement produces the next focused
query.  Retrieval stops when all requirements are covered, four rounds are
exhausted, or a duplicate follow-up cannot be replaced.  The same 0.6B Qwen3
pointwise model scores the accumulated candidates once.  We retain its top 60
for a capacity-matched comparison with the baseline and expose up to its top
120 to the operational set selector.  If checklist parsing fails, the system
performs one search with the original question; if a later audit cannot be
parsed, it stops expansion while retaining all candidates accumulated so far.

\subsection{Requirement-seeded setwise selection}
The pointwise control returns the ten highest relevance scores.  Our selector
instead seeds a ten-item set with the highest-ranked cited turn for each covered
requirement, then fills empty slots by the pointwise prior.  The first five
positions of this \emph{seeded set} are protected; they are not necessarily the
original pointwise top five.  Remaining candidates from the top 120 arrive in
batches of eight.  Qwen3-32B may replace at most one item per batch and three
items overall, and only when the challenger directly adds or more strongly
supports a listed requirement while the removed item is redundant.  Invalid
JSON, unknown identifiers, or an unsupported justification leave the set
unchanged.  Pointwise and setwise contexts therefore share $A_q$, the same
top-120 candidate set, pointwise scores, and $k=10$; only the final set utility differs.

Figure~\ref{fig:method} shows the complete intervention.  Gold support turns are
used only after retrieval and generation for evaluation.

\section{Experimental Setup}
\textbf{Data.}  The evaluated LoCoMo release contains ten generated,
human-verified conversations: 5,882 turns (588.2 per conversation), up to
32 sessions, and 1,986 QA examples \citep{maharana2024locomo}.
REALTALK contains ten
21-day conversations between people with 8,944 turns and 728 memory-probing
questions: 301 factual, 319 temporal, and 108 multi-hop
\citep{lee2025realtalk}.  The multi-hop slice has 327 annotated support turns.

\textbf{Models and budgets.}  Both controlled multi-hop systems use the same
BM25 index, E5-Mistral-7B-Instruct embeddings, RRF weights, 0.6B Qwen3 pointwise
scorer (maximum input length 4,096), and top-10 reader context.  The flat
baseline stores 60 pointwise-ranked candidates.  Requirement-guided discovery
reports a matched top-60 candidate set and an operational top-120 prefilter; the latter
is shared by its pointwise and setwise top-10 variants.  The requirement planner
and setwise selector use Qwen3-32B with temperature zero and thinking disabled
\citep{yang2025qwen3}.  The 0.6B pointwise scorer is non-generative and runs in
evaluation mode without gradients.  The frozen reader is
\texttt{gpt-oss-120b} at temperature zero \citep{openai2025gptoss}.  The full
baseline evaluation allows 6,000 context tokens; saved multi-hop contexts are
replayed with a 4,000-token cap and no new retrieval.  Qwen3-32B judge calls also use
temperature zero.

\textbf{Generation and citations.}  Query-type-specific prompts ask for concise
factual, temporal, or multi-hop answers.  Citation repair is gold-blind and can
choose only a packed turn.  It preserves an already valid citation; otherwise,
for packed turn $t$ at rank $j$, it computes
\begin{equation}
  3|W(y_q)\cap W(t)|+0.5|W(q)\cap W(t)|+1/j,
\end{equation}
where $W$ removes common weak tokens, then cites up to two source identifiers
from the highest-scoring turn.  It cannot introduce unretrieved evidence.

\textbf{Protocol and reproducibility.}  Principal tables use the frozen
baseline and EviCover runs; three additional full retrieval replications
measure stability under the same configuration.  Temperature zero removes
intentional sampling, not all remote-serving nondeterminism.  Principal
multi-hop comparisons use all 108 questions.  A repeatedly
inspected fixed first-30 slice rejects expensive alternatives and is diagnostic,
not held-out.  The full-108 analysis includes these development questions;
it is an in-corpus audit, not an independent held-out confirmation.
Each run stores configuration, examples, retrieval traces, token
use, packed contexts, metrics, and resumable state under a UUID.  Paper tables
and confidence intervals are regenerated from immutable JSONL artifacts.

\section{Results}
\subsection{RQ1: Dataset evaluation and stage-wise proof-set loss}
Table~\ref{tab:datasets} shows that revising answer-generation settings on LoCoMo
improves token F1 and citation-grounded answers while Recall@5 remains nearly constant.
With the revised configuration held fixed, REALTALK results are lower on all
reported answer and citation measures.  The two datasets have different question
distributions, so these scores do not establish their relative difficulty.

\begin{table}[t]
\caption{Baseline evaluation on both datasets. CP/CR are citation precision/recall;
Cit.-Gr. is the mechanical citation-grounded answer rate defined in Sec.~\ref{sec:metrics}.}
\label{tab:datasets}
\centering
\footnotesize
\setlength{\tabcolsep}{3.3pt}
\begin{tabular}{lccccc}
\toprule
Run & R@5 & F1 & CP & CR & Cit.-Gr. \\
\midrule
LoCoMo initial & .772 & .316 & .463 & .448 & .493 \\
LoCoMo revised & .771 & \textbf{.394} & \textbf{.558} & \textbf{.535} & \textbf{.593} \\
REALTALK baseline & .579 & .291 & .477 & .374 & .511 \\
\bottomrule
\end{tabular}
\end{table}

The LLM judge assigns higher semantic scores than lexical metrics suggest:
on all 728 questions, 55.9\% are accepted (including accepted abstentions), 80.2\%
are accepted or partial, and 71.6\% are supported by the shown context, compared
with 6.7\% EM.  Table~\ref{tab:slices} nevertheless identifies multi-hop as the
weakest slice for both retrieval and judge support.  Temporal F1 is additionally
affected by equivalent date formats and concise surface forms.

\begin{table}[t]
\caption{REALTALK baseline by query type. J-C/P is judge correct-or-partial;
J-Sup. is judge-supported; Cit.-Gr. is the mechanical metric.}
\label{tab:slices}
\centering
\footnotesize
\setlength{\tabcolsep}{4.0pt}
\begin{tabular}{lrrrrr}
\toprule
Type ($n$) & R@5 & F1 & J-C/P & J-Sup. & Cit.-Gr. \\
\midrule
Factual (301) & .391 & .381 & .847 & .787 & .475 \\
Temporal (319) & \textbf{.814} & .239 & .777 & .702 & \textbf{.611} \\
Multi-hop (108) & .409 & .191 & .750 & .556 & .315 \\
\bottomrule
\end{tabular}
\end{table}

\textbf{Temporal and semantic distance.}  For each annotated support identifier,
we measure its distance from the final released turn of the same conversation.
This is a lower bound on distance to the memory-probing question if that
question was asked later.  Table~\ref{tab:distance} confirms that the task is
not merely short-range semantic retrieval: the median support turn is
\EvidenceDayLagMedian{} days and \EvidenceTurnLagMedian{} turn positions old,
and \EvidenceAtLeastSevenDays{} of all support turns are at least seven days
old.  At the question level, the oldest required support has a median age of
\OldestEvidenceDayMedian{} days.

\begin{table}[t]
\caption{Distance from each of 327 support turns to the final observed turn.}
\label{tab:distance}
\centering
\footnotesize
\setlength{\tabcolsep}{7.0pt}
\begin{tabular}{lrrr}
\toprule
Distance unit & Median & P90 & Maximum \\
\midrule
Turn positions & \EvidenceTurnLagMedian{} & \EvidenceTurnLagPninety{} & \EvidenceTurnLagMax{} \\
Sessions & \EvidenceSessionLagMedian{} & \EvidenceSessionLagPninety{} & \EvidenceSessionLagMax{} \\
Days & \EvidenceDayLagMedian{} & \EvidenceDayLagPninety{} & \EvidenceDayLagMax{} \\
\bottomrule
\end{tabular}
\end{table}

Temporal distance is not the only difficulty.  Among the
\EvidenceTextAvailable{} support turns present as text in the release,
\EvidenceZeroLexicalOverlap{} have zero content-word overlap with their
question.  We lowercase, retain alphanumeric tokens longer than one character,
and remove the fixed English stopword list supplied with the diagnostic code.
This is a surface-form proxy, not a semantic similarity measure.
The remaining \EvidenceMediaOnly{} annotated identifiers refer to
media-only turns absent from the textual export; they are retained in temporal
counts but excluded from lexical-overlap statistics.  Recall by age is not
monotonic: baseline final ER is .324 for 0--2 days, .407 for 3--6, .322 for
7--13, and .349 for 14+ days.  The benchmark therefore combines genuine
long-range memory with semantic mismatch rather than isolating either factor.

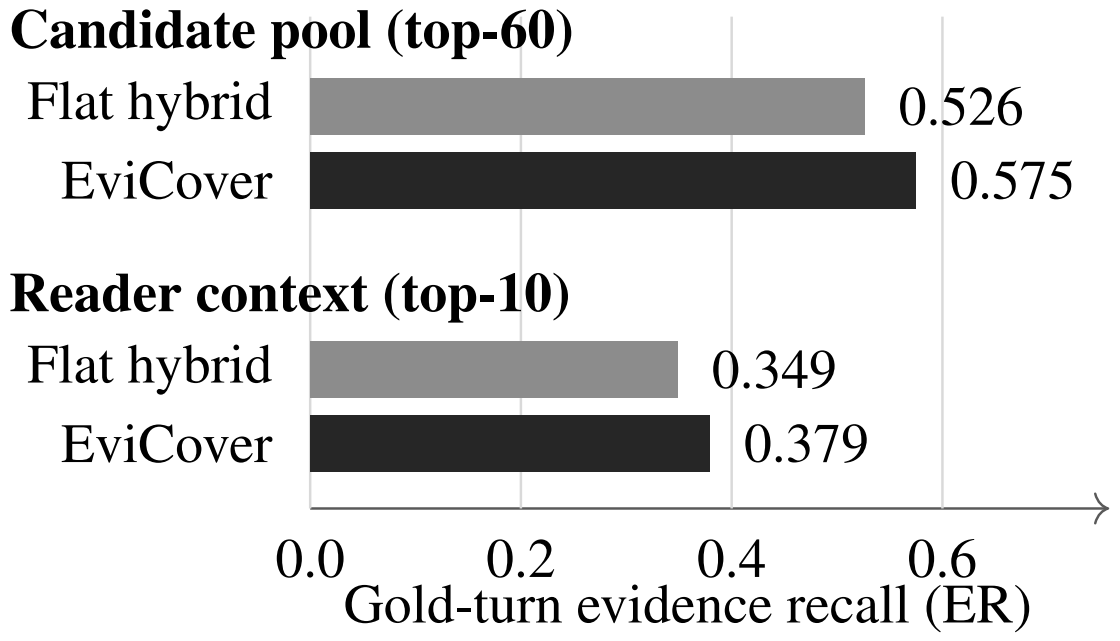
\begin{figure}[t]
  \centering
  \resizebox{0.98\columnwidth}{!}{\input{figures/evidence_funnel.tex}}
  \caption{Annotated support-turn evidence recall. Candidate bars compare matched
  top-60 budgets; final-context bars compare top-10 outputs.}
  \label{fig:funnel}
\end{figure}

Table~\ref{tab:funnel} and Fig.~\ref{fig:funnel} trace all 327 multi-hop support
turns.  The flat baseline retains 172 in its 60 pointwise-ranked candidates but only
114 in final context.  Requirement-guided retrieval accumulates 217 and retains
188 in its matched top-60 candidate set: $+0.049$ ER with a 95\%
conversation-cluster bootstrap CI of $[0.028,0.067]$.  Allowing the operational
prefilter to contain 120 candidates retains 207 turns; this number measures
available evidence for set selection and is not a capacity-matched baseline
comparison.  Setwise final context retains 124 turns, $+0.031$ over the complete
flat pipeline (CI $[0.007,0.064]$).

\begin{table}[t]
\caption{Evidence funnel on all 108 multi-hop questions. ``Req. discovery''
denotes requirement-guided retrieval. Top-60/120 rows report ranked candidates.}
\label{tab:funnel}
\centering
\footnotesize
\setlength{\tabcolsep}{3.8pt}
\begin{tabular}{llrrr}
\toprule
System & Stage & Hits/327 & ER & Full sets \\
\midrule
Flat hybrid & top-60 & 172 & .526 & 50/108 \\
Flat hybrid & final top-10 & 114 & .349 & 36/108 \\
\midrule
Req. discovery & accumulated & 217 & .664 & 63/108 \\
Req. discovery & top-60 & 188 & .575 & 55/108 \\
Req. discovery & top-120 & 207 & .633 & 60/108 \\
Pointwise & final top-10 & 116 & .355 & 36/108 \\
Setwise & final top-10 & \textbf{124} & \textbf{.379} & \textbf{38/108} \\
\bottomrule
\end{tabular}
\end{table}

Setwise selection improves ER over the matched pointwise top-10 from .355 to .379
($+0.024$, CI $[0.009,0.047]$), but full proof sets increase only from 36 to 38;
the $+0.019$ rate difference has CI $[-0.018,0.055]$.  An oracle top-10 from the
same candidate set reaches .624 ER and preserves all 60 annotated support sets
available in full.  The 108
questions decompose into 45 whose complete set is never discovered, three lost
during independent scoring, 22 complete among ranked candidates but incomplete
after top-10 selection, and 38 complete in reader context.

Exploratory alternatives on the fixed-30 slice and additional memory
representations are reported in Appendix~\ref{app:alternatives}.

\subsection{RQ2: Evidence gains do not yet establish answer gains}
We replay answers over the 108 saved setwise contexts with the same reader,
query-type prompt family, temperature, and citation policy.  No retrieval is
repeated.  Table~\ref{tab:answers} shows that judge accepted-or-partial and
supported rates each rise by .028, but both confidence intervals include zero.
Mechanical F1 and citation-grounded rate decline slightly, also with intervals
crossing zero.  Correct labels change from 35 to 32, correct abstentions from 31
to 35, partial answers from 15 to 17, and wrong abstentions from 9 to 6.  The new
context therefore changes answer behavior without a statistically clear gain in
full correctness.

\begin{table}[t]
\caption{Paired answer replay on 108 multi-hop questions. CIs are for $\Delta$.}
\label{tab:answers}
\centering
\footnotesize
\setlength{\tabcolsep}{3.0pt}
\begin{tabular}{lrrrr}
\toprule
Metric & Flat & Req.+set & $\Delta$ & 95\% CI \\
\midrule
Judge accepted & .611 & .620 & +.009 & [-.058,.074] \\
Accepted/partial & .750 & .778 & +.028 & [-.020,.083] \\
Judge supported & .556 & .583 & +.028 & [-.037,.080] \\
Judge score & 3.083 & 3.028 & -.056 & [-.180,.065] \\
Token F1 & .191 & .183 & -.009 & [-.023,.006] \\
Citation-grounded & .315 & .306 & -.009 & [-.063,.057] \\
\bottomrule
\end{tabular}
\end{table}

Complete evidence remains associated with answer quality: when the setwise
context contains the full annotated proof set, judge correct-or-partial is .895
and support is .737, versus .714 and .500 otherwise.  This observational gap is
not causal because easier questions may also be easier to retrieve, but it
supports reporting complete-set survival rather than only the first relevant
hit.

\subsection{Reader-only intervention on frozen contexts}
We next change only the multi-hop reader instruction on the same 108 saved
EviCover contexts.  The original prompt favors minimal answer spans and omits
reasoning; the revised prompt permits one compact, explicitly supported reason,
requires citations for its evidence, and prohibits unsupported elaboration.
The model, temperature, 4,000-token context cap, and citation repair remain
unchanged.  This is a post-hoc reader ablation, not an independent test set or
an additional retrieval improvement.

Table~\ref{tab:reader} reports the completed paired replay; all 108 judge
responses parse successfully.  Substantively correct labels increase from
32 to 36, but accepted abstentions decrease from 35 to 24 and wrong abstentions
increase from 6 to 10.  Partial and incorrect labels change from 17 to 21 and
18 to 17, respectively.  Consequently, the overall accepted rate falls from
.620 to .556 despite the higher mean score and citation-grounded rate.
Judge-supported answers change only from 63 to 65.  The accepted-rate decrease
has a conversation-cluster CI of $[-.131,-.009]$, while the citation-grounded
increase has CI $[.010,.126]$; these exploratory intervals are not adjusted for
multiple comparisons.  Better citation overlap
therefore does not establish better semantic correctness; we retain the
original reader for the principal EviCover comparison.

\begin{table}[t]
\caption{Reader-only paired ablation on identical EviCover contexts (108
questions). Accepted includes judge-accepted abstentions.}
\label{tab:reader}
\centering
\footnotesize
\setlength{\tabcolsep}{2.5pt}
\input{stats/reader_comparison.tex}
\end{table}

\subsection{Temperature-zero repeatability}
Three independent replications use the same 108 questions, indexes, budgets,
prompts, and decoding configuration.  Across pairs, the final selected
identifiers match exactly on 91.7--99.1\% of questions; mean evidence-set
Jaccard similarity is .971--.998.  Recall@5 ranges from .4003 to .4020
(spread .0017), MRR from .4664 to .4667, and nDCG@5 from .3814 to .3838.
Thus aggregate retrieval performance is stable in these replications, but the
generated search trajectories and individual contexts are not identical.
These are retrieval repetitions, not repeated answer generation or independent
draws of new conversations.

\subsection{Efficiency audit}
Table~\ref{tab:cost} reports exact saved-run cost on the same 108 questions.
Mean retrieval latency increases by \EviCoverLatencyRatio{}$\times$.  EviCover
adds \EviCoverCallsPerQuery{} Qwen3-32B calls and
\EviCoverTokensPerQuery{} planner/selector tokens per question, for
\EviCoverTotalTokens{} tokens in total; \EviCoverSelectorTokenShare{} are spent
by final set selection.  Relative to the flat final context, this corresponds
to \EviCoverTokensPerAddedEvidence{} tokens per additional retained support turn
and \EviCoverTokensPerAddedProofSet{} per additional complete proof set.  These
are inference counts, not monetary or energy estimates, because both models ran
on private endpoints without hardware accounting; the wall-clock ratio is
observational rather than hardware-normalized.

\begin{table}[t]
\caption{Retrieval utility and cost on the same 108 multi-hop questions.  LLM
calls/tokens count additional generative Qwen3-32B retrieval work.}
\label{tab:cost}
\centering
\footnotesize
\setlength{\tabcolsep}{4.0pt}
\begin{tabular}{lrr}
\toprule
Measure & Flat hybrid & EviCover \\
\midrule
Mean latency (s/query) & \BaselineLatencyMean{} & \EviCoverLatencyMean{} \\
Pointwise candidates/query & \BaselinePointwisePoolMean{} & \EviCoverPointwisePoolMean{} \\
Generative calls/query & 0 & \EviCoverCallsPerQuery{} \\
Generative tokens/query & 0 & \EviCoverTokensPerQuery{} \\
Final ER & \BaselineFinalRecall{} & \SetwiseFinalRecall{} \\
Complete proof sets & \BaselineCompleteSets{} & \SetwiseCompleteSets{} \\
\bottomrule
\end{tabular}
\end{table}

\section{Discussion}
\textbf{Item relevance and set coverage are different objectives.}  The 22
questions with complete support among ranked candidates but incomplete support
in the final context quantify the mismatch.  A pointwise
ranker can repeatedly select turns about the same aspect, whereas proof-set
selection values the marginal coverage of a currently missing requirement.
The oracle gap shows that a learned set utility remains a larger opportunity
than further hand-tuning RRF weights.

\textbf{Human dialogue changes the search target.}  LoCoMo multi-hop
questions often combine explicit facts across time.  REALTALK also asks for
judgments and preferences inferred from behavior; the abstract term in the
question may be absent from its support turns.  Observable requirements improve
discovery under this lexical mismatch, but Qwen3-generated requirements remain
imperfect and expensive.

\textbf{Reader utilization is a separate bottleneck.}  Better proof-set recall
mostly changes abstention and partial-credit labels rather than full correctness.
Future models should jointly score requirement coverage, provenance, redundancy,
and whether the reader can compose an answer from the selected set.

\textbf{The intervention is diagnostic, not yet operationally efficient.}
Most added computation is spent repeatedly deciding whether a candidate adds
marginal coverage, yet only two additional complete proof sets reach the reader
and answer intervals cross zero.  EviCover is therefore evidence that pointwise
selection is a real bottleneck, not evidence that this particular 32B
implementation should be a universal production default.  A practical successor
would distill set utility into a smaller scorer or invoke it only when an
inexpensive coverage test detects a likely incomplete proof set.

\section{Conclusion}
Stage-wise analysis reveals two separable failures in long-term conversational
memory: complete proof sets are often never discovered, and complementary turns
that are discovered are discarded before generation.  Provenance-constrained
requirement discovery improves the REALTALK candidate set at matched capacity,
and setwise selection preserves a small portion of that gain, but at substantial
inference cost.  Complete-set and answer-level improvements remain modest and
uncertain.  Reporting temporal and lexical distance together with ER and PS
through the retrieval funnel gives a more actionable account than a single
top-$k$ or answer score: better memory requires complete evidence delivery, an
efficient set objective, and a reader able to use the delivered evidence.

\section*{Limitations}
The principal comparison has 108 questions nested in ten conversations.
Conversation-cluster bootstrap respects that dependence, but ten clusters yield
wide intervals.  The fixed-30 slice was repeatedly inspected, and exploratory
alternatives and the later reader-prompt change are descriptive rather than
confirmatory.  Intervals are conditional on the selected configurations, not
adjusted for method selection or multiple exploratory comparisons.
Retrieval replications measure inference stability on the same
data, not generalization or repeatability of the answer evaluation.
We use one reader model and one uncalibrated LLM judge.  The judge sees
mechanical scores and gold identifiers, is not blinded to those signals, and
uses the same Qwen3-32B model family as discovery and selection.  Its
accepted-abstention labels must not be equated with corpus-level
unanswerability; no human adjudication of those labels was performed.

LoCoMo and REALTALK have different question distributions, so cross-dataset scores
do not rank intrinsic dataset difficulty.  Temporal lag is measured to the
final released turn rather than a question timestamp; it does not measure
the number of turns given to the reader.  Thirty-three media-only support
identifiers lack released text.  Annotated support may also be incomplete,
especially for inferential questions; PS measures survival of the released
annotation, not every valid proof.  The controlled alternatives use
implementations inspired by published approaches, not their trained
checkpoints or full published training recipes, and do not establish
leaderboard comparisons with those systems.

This study evaluates memory-question answering, not personalization: it does
not test whether an assistant adapts recommendations, style, or actions to an
individual user.  Long-term memory can support personalization, but the present
results should not be interpreted as evidence of personalized behavior.
Hardware-normalized latency, monetary cost, and energy use were not measured.

\section*{Ethical considerations}
REALTALK contains conversations between people.  We use only the released benchmark,
add no human-subject data, preserve source provenance, and do not quote source
conversations in the manuscript.  Illustrative examples are schematic rather than
verbatim dialogue.  A deployed memory system would require access
control, deletion, retention limits, and user consent.  We do not redistribute
conversation text as part of the manuscript package; access remains governed
by the source datasets' terms.

\section*{Acknowledgment}
This work was supported by the Ministry of Economic Development of the Russian
Federation (IGK 000000C313925P4C0002), agreement No 139-152025-010.

\bibliography{references}
\appendix
\section{Exploratory alternatives}
\label{app:alternatives}
Table~\ref{tab:pilots} summarizes the fixed-30 diagnostic slice (82 support
turns).  Free iterative search, query-local graph chaining, breadth-then-depth
expansion, and a restricted executable search program broaden discovery, but
none exceeds requirement-guided support retention after the shared pointwise
reranker.  The repeated gap between accumulated and final ER motivated the
setwise intervention rather than another query-expansion prompt.

\begin{table}[htbp]
\caption{Exploratory fixed-30 multi-hop diagnostics (82 support turns).}
\label{tab:pilots}
\centering
\footnotesize
\setlength{\tabcolsep}{3.5pt}
\begin{tabular}{lrrrr}
\toprule
Discovery method & R@5 & \shortstack{Accum.\\ER} & \shortstack{Ranked\\ER} & \shortstack{Final\\ER} \\
\midrule
Flat lexical+dense & \textbf{.452} & -- & .524 & \textbf{.378} \\
Free iterative & .397 & .646 & .549 & .366 \\
Requirement-guided & .424 & \textbf{.683} & \textbf{.573} & \textbf{.378} \\
\shortstack[l]{Query-local graph\\chain} & .374 & .659 & .512 & .329 \\
Breadth-then-depth & .358 & .671 & .549 & .329 \\
\shortstack[l]{Restricted search\\program} & .386 & .634 & .537 & .341 \\
\bottomrule
\end{tabular}
\end{table}

Alternative representations expose the same boundary.  A proposition index
extracts 14,563 short facts from 8,925 processed turns and raises full multi-hop
candidate hits from 172 to 188, but final hits change only from 114 to 115; a
proposition-aware selector reduces them to 91.  A provenance-linked index of
2,798 episodic events finds seven support turns absent from the baseline candidate set on a
30-question behavioral subset, but quota fusion does not retain them reliably.
A ready-made ColBERT-style model matches baseline candidate ER (.524) yet lowers final ER
(.341 versus .378).  The proposition and event controls show that richer
representations can surface complementary turns, but their gains are not
automatically preserved by the shared itemwise objective.

\end{document}

%% file: stats/paper_statistics.tex
\newcommand{\BaselinePoolRecall}{0.526}
\newcommand{\RequirementPoolMatchedRecall}{0.575}

\newcommand{\BaselineFinalRecall}{0.349}
\newcommand{\SetwiseFinalRecall}{0.379}
\newcommand{\BaselineCompleteSets}{36}
\newcommand{\SetwiseCompleteSets}{38}

\newcommand{\EvidenceTurnLagMedian}{380}
\newcommand{\EvidenceTurnLagPninety}{989}
\newcommand{\EvidenceTurnLagMax}{1574}
\newcommand{\EvidenceSessionLagMedian}{12}
\newcommand{\EvidenceSessionLagPninety}{19}
\newcommand{\EvidenceSessionLagMax}{24}
\newcommand{\EvidenceDayLagMedian}{12}
\newcommand{\EvidenceDayLagPninety}{21}
\newcommand{\EvidenceDayLagMax}{23}
\newcommand{\OldestEvidenceDayMedian}{16.5}
\newcommand{\EvidenceAtLeastSevenDays}{72.2\%}

\newcommand{\EvidenceZeroLexicalOverlap}{63.3\%}
\newcommand{\EvidenceTextAvailable}{294}
\newcommand{\EvidenceMediaOnly}{33}
\newcommand{\BaselineLatencyMean}{18.28}
\newcommand{\EviCoverLatencyMean}{41.62}
\newcommand{\EviCoverLatencyRatio}{2.28}
\newcommand{\EviCoverCallsPerQuery}{17.47}
\newcommand{\EviCoverTokensPerQuery}{24.59K}
\newcommand{\EviCoverTotalTokens}{2.66M}
\newcommand{\EviCoverSelectorTokenShare}{81.2\%}
\newcommand{\BaselinePointwisePoolMean}{60}
\newcommand{\EviCoverPointwisePoolMean}{118.45}
\newcommand{\EviCoverTokensPerAddedEvidence}{265,568}
\newcommand{\EviCoverTokensPerAddedProofSet}{1,327,838}

%% file: figures/method_overview.tex
\begin{tikzpicture}[x=1cm,y=1cm,font=\footnotesize,
  main/.style={draw=black!70,align=center,text width=3.25cm,
    minimum height=1.0cm,inner sep=4pt,fill=black!3},
  arrow/.style={-{Latex[length=2mm]},line width=.6pt}]
\node[main] (requirements) at (0,0) {Question $q$\\Qwen3-32B: 1--4\\observable requirements};
\node[main] (search) at (4,0) {BM25 + E5-Mistral\\search for uncovered\\requirements};
\node[main] (audit) at (8,0) {Qwen3-32B coverage audit\\must cite retrieved IDs};
\node[main] (accumulated) at (12,0) {Unique discovered turns\\$A_q$: measure ER, PS};
\node[main] (pool) at (12,-1.9) {Qwen3-Reranker-0.6B\\scores turns independently\\$P_q$: measure ER, PS};
\node[main] (pointwise) at (7.8,-1.9) {Pointwise top-10\\control context};
\node[main] (setwise) at (7.8,-3.5) {Requirement-seeded\\setwise top-10 context};
\node[main] (reader) at (3.6,-2.7) {Fixed reader: gpt-oss-120b\\context $C_q$: ER, PS\\answers and citations};
\draw[arrow] (requirements) -- (search);
\draw[arrow] (search) -- (audit);
\draw[arrow] (audit) -- (accumulated);
\draw[arrow] (accumulated) -- (pool);
\draw[arrow] (pool) -- (pointwise);
\draw[arrow] (pool.south) |- (setwise.east);
\draw[arrow] (pointwise.west) -- ([yshift=3mm]reader.east);
\draw[arrow] (setwise.west) -- ([yshift=-3mm]reader.east);
\draw[arrow,dashed] (audit.north) -- ++(0,.65) -| (search.north)
  node[pos=.25,above] {uncovered: next round (at most 4)};
\end{tikzpicture}

%% file: figures/evidence_funnel.tex
\begin{tikzpicture}[x=1cm,y=1cm,font=\footnotesize]
  \draw[->,black!65] (0,0) -- (4.55,0);
  \draw[black!15] (0.000,0) -- (0.000,2.8);
  \node[anchor=north] at (0.000,-0.04) {0.0};
  \draw[black!15] (1.200,0) -- (1.200,2.8);
  \node[anchor=north] at (1.200,-0.04) {0.2};
  \draw[black!15] (2.400,0) -- (2.400,2.8);
  \node[anchor=north] at (2.400,-0.04) {0.4};
  \draw[black!15] (3.600,0) -- (3.600,2.8);
  \node[anchor=north] at (3.600,-0.04) {0.6};
  \node[anchor=west,font=\bfseries\footnotesize] at (-1.85,2.700) {Candidate pool (top-60)};
  \node[anchor=east] at (-.08,2.290) {Flat hybrid};
  \node[anchor=east] at (-.08,1.870) {EviCover};
  \fill[black!45] (0,2.130) rectangle (3.156,2.450);
  \fill[black!85] (0,1.710) rectangle (3.450,2.030);
  \node[anchor=west] at (3.216,2.290) {0.526};
  \node[anchor=west] at (3.510,1.870) {0.575};
  \node[anchor=west,font=\bfseries\footnotesize] at (-1.85,1.200) {Reader context (top-10)};
  \node[anchor=east] at (-.08,0.790) {Flat hybrid};
  \node[anchor=east] at (-.08,0.370) {EviCover};
  \fill[black!45] (0,0.630) rectangle (2.092,0.950);
  \fill[black!85] (0,0.210) rectangle (2.275,0.530);
  \node[anchor=west] at (2.152,0.790) {0.349};
  \node[anchor=west] at (2.335,0.370) {0.379};
  \node at (2.15,-.57) {Gold-turn evidence recall (ER)};
\end{tikzpicture}

%% file: stats/reader_comparison.tex
\begin{tabular}{lrrrr}
\toprule
Metric & Original & Revised & $\Delta$ & 95\% CI \\
\midrule
Judge accepted & .620 & .556 & -.065 & [-.131,-.009] \\
Accepted/partial & .778 & .750 & -.028 & [-.095,.028] \\
Judge supported & .583 & .602 & +.019 & [-.029,.071] \\
Judge score & 3.028 & 3.157 & +.130 & [-.020,.284] \\
Token F1 & .183 & .147 & -.036 & [-.111,.031] \\
Citation-grounded & .306 & .370 & +.065 & [.010,.126] \\
\bottomrule
\end{tabular}